\documentclass[10pt,twocolumn,letterpaper]{article}
\usepackage{textcomp}  
\usepackage[symbol]{footmisc}

\usepackage{wacv}              

\usepackage{tabularx}
\usepackage{makecell}
\usepackage{float}
\usepackage{cuted}
\usepackage{caption}

\usepackage[normalem]{ulem} 
\usepackage{xcolor}

\newcommand{\spar}[1]{\smallskip\noindent\textbf{#1}}

\usepackage{tabularx,booktabs,makecell,xcolor}
\usepackage{siunitx}
\definecolor{wacvblue}{rgb}{0.21,0.49,0.74}
\usepackage[pagebackref,breaklinks,colorlinks,allcolors=wacvblue]{hyperref}

\def\wacvPaperID{3019} 
\def\confName{WACV}
\def\confYear{2026}

\title{FAST-EQA: Efficient Embodied Question Answering with Global and Local Region Relevancy}

\usepackage[symbol]{footmisc}
\usepackage{hyperref}

\author{
Haochen Zhang$^{1}$\textsuperscript{†}\thanks{Equal contribution. \textsuperscript{†}Work done during internship at HRI.}\hspace{1cm}
Nirav Savaliya$^{2}$\footnotemark[1] \hspace{1cm}
Faizan Siddiqui$^{2}$ \hspace{1cm}
Enna Sachdeva$^{2}$
\\
\\
$^{1}$Carnegie Mellon University, USA
\hspace{1cm}$^{2}$Honda Research Institute USA \\
{\tt\small haochen4@andrew.cmu.edu}, 
{\tt\small \{nsavaliya, faizan\_siddiqui, enna\_sachdeva\}@honda-ri.com}
}

\begin{document}
\maketitle

\vspace*{-3em}
\begin{strip}
    \centering
     \includegraphics[width=1.0\textwidth]{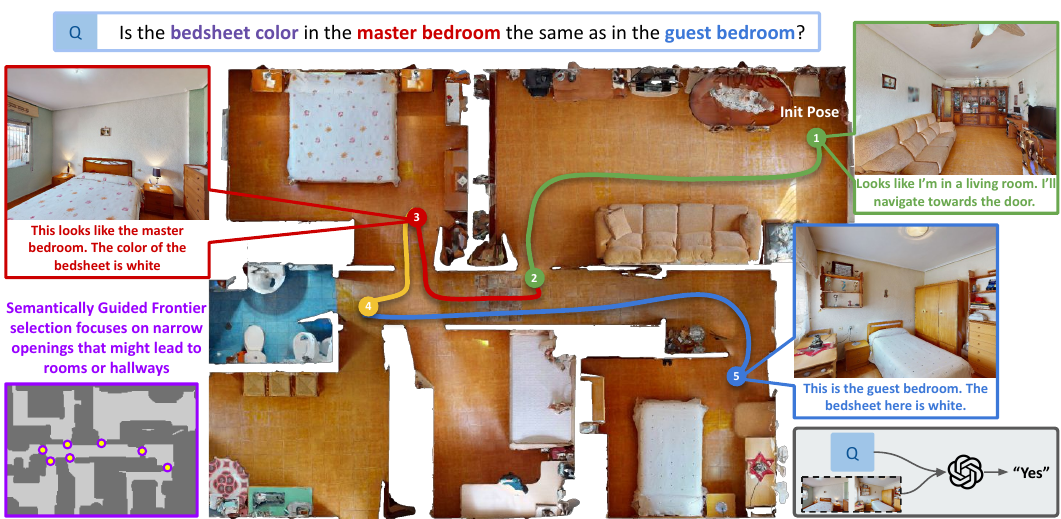}
    \captionof{figure}{In the illustrated scenario, our FAST-EQA agent first localizes relevant regions, such as the master bedroom and guest bedroom, and identifies the visual target: bedsheet. Guided by a semantic-aware global exploration strategy focused on relevant rooms, it navigates across these regions while maintaining and updating a target-specific memory based on visual relevance. Once sufficiently confident, the agent queries a large vision–language model (here, GPT-4o) to answer the question using the stored visual observations.}
    \label{fig:overview}
\end{strip}

\begin{abstract}

Embodied Question Answering (EQA) combines visual scene understanding, goal‑directed exploration, spatial and temporal reasoning under partial observability. A central challenge is to confine physical search to question‑relevant subspaces while maintaining a compact, actionable memory of observations. Furthermore, for real-world deployment, fast inference time during exploration is crucial. We introduce FAST‑EQA, a question‑conditioned framework that (i) identifies likely visual targets, (ii) scores global regions of interest to guide navigation, and (iii) employs Chain‑of‑Thought (CoT) reasoning over visual memory to answer confidently. FAST‑EQA maintains a bounded scene memory that stores a fixed‑capacity set of region–target hypotheses and updates them online, enabling robust handling of both single‑ and multi‑target questions without unbounded growth. To expand coverage efficiently, a global exploration policy treats narrow openings and doors as high‑value frontiers, complementing local target seeking with minimal computation. Together, these components focus the agent’s attention, improves scene coverage, and improve answer reliability while running substantially faster than prior approaches. On HMEQA and EXPRESS‑Bench, FAST‑EQA achieves state‑of‑the‑art performance, while performing competitively on OpenEQA and MT‑HM3D. All source code will be publicly released. More information can be found on our project page: \href{https://astronirav.github.io/fasteqa}{https://astronirav.github.io/fasteqa}

\end{abstract}
    
\section{Introduction}
\label{sec:intro}

In order to have generalizable robot assistants in human-centric environments such as homes, we seek the ability for robots to actively explore unknown environments and handle diverse natural language queries. This behavior can be characterized by the task of Embodied Question Answering (EQA), where a physical or simulated embodied agent is asked to answer a natural‑language question via moving and perceiving in an unexplored 3D environment, rather than relying on a static image. The agent must combine natural language understanding, visual scene understanding, exploration, and spatial and temporal reasoning under partial observability to gather the evidence needed to answer. A core challenge is deciding where to explore at each step, narrowing the search to question‑relevant regions while keeping a compact memory representation of what has been seen, especially in unseen environments. To add more complexity, practical tasks often have multiple targets, requiring embodied agent to navigate and gather information from multiple sources to answer the question. For instance, to answer the question \textit{“Are the curtains in the bedroom the same color as those in the living room?”}, the agent must visit both the rooms, detect the curtains in each, store the observations in memory, and then reason over this evidence to provide the answer. Finally, in order to support physical robots in the real world, EQA approaches must follow deployment constraints with regards to power, memory, and inference time.

Recent advances in large language models (LLMs) and vision-language models (VLMs) have shown impressive capabilities in planning, reasoning, and question-answering (QA) \cite{zhang2024navid, li2023blip, liu2023visual} in the field of 2D vision. Thus, existing EQA systems increasingly lean on these models to choose exploration directions, quantify confidence, and inject pre-trained, web-scale knowledge that can anchor high‑level goals. However, VLMs struggle with long-horizon embodied AI tasks due to shallow working memory, limited spatial reasoning, and weak integration of observations over time. Trained primarily on static image–text corpora, they lack the sequential perception–action grounding needed to maintain coherent representations of large, partially observed scenes or to selectively retain task-relevant information \cite{wang2025embodied}. These issues become especially acute in multi-target tasks, where agents must jointly reason over spatial and temporal evidence gathered from multiple locations. 

To mitigate these limitations, many methods maintain structured memory in the form of object-centric metric-semantic scene graphs \cite{Hydra, HOVSG, OSG}. While effective for capturing symbolic relationships, explicit graph construction is computationally and memory intensive, slows real-time operation, and often collapses nuanced spatial detail into coarse edges. Alternatively, storing raw images or memory snapshots \cite{3DMem} preserves fine-grained layout but causes memory usage to grow without bound as exploration proceeds—an issue that is particularly acute in long-horizon and multi-target tasks. Thus, existing memory designs trade off either efficiency or scalability, limiting their practical utility.

Furthermore, exploration strategies present parallel challenges. Most existing approaches are frontier-based \cite{HMEQA}, which are effective in open spaces but blind to indoor regularities, inattentive to question‑conditioned semantics, prone to repeated exploration when uncertain, and unable to target intuitive transitions such as doors, hallways, or rooms that are often key to exploration. Consequently, both memory and control components of existing EQA systems struggle to scale gracefully to complex, long-horizon tasks.

In the context of the above challenges, we present FAST‑EQA - FAst, Semantics-aware, Target-driven Exploration for Embodied Question Answering (Figure \ref{fig:overview}), an embodied QA system that couples semantically-guided global and local exploration policies with bounded visual memory. Given a question, FAST‑EQA first uses an LLM to extract candidate visual goals or targets, initializes a spatial memory and ranks regions of the environment by their likelihood of containing relevant evidence for the goals. Exploration then proceeds with two complementary policies: Global Relevance (GR) Exploration, which departs from standard frontier-based exploration (FBE) by prioritizing transitional waypoints such as hallways and doors to efficiently reach promising areas, and Local Relevance (LR) Exploration, which assess relevant local regions (e.g. rooms) on their informativeness for answering the question. The agent interleaves GR and LR steps dynamically until it can answer confidently. As the agent explores, it maintains an explicitly-bounded visual memory, maintaining a fixed budget of $k$ visual snapshots per target to ensure efficiency and scalability. After taking LR steps, the system invokes the VLM’s chain‑of‑thought reasoning over these snapshots to produce the final answer.

We evaluate FAST‑EQA comprehensively on both multiple-choice QA and open-answer datasets across HM-EQA \cite{HMEQA}, EXPRESS‑Bench \cite{FineEQA}, OpenEQA \cite{OpenEQA}, and MT‑HM3D \cite{MemoryEQA}, achieving superior performance on HM-EQA and EXPRESS‑Bench and competitive results on OpenEQA and MT‑HM3D, underscoring the method’s versatility across question types. Additionally, we show that our method has superior real-time inference capabilities, consistently producing navigation decisions faster during each step of exploration. Our contributions can be summarized as follows:

\begin{itemize}
  \item We present FAST-EQA, a lightweight embodied QA framework that runs in near real-time with a compact memory footprint, making it well-suited for deployment on embodied agents.
  \item We propose semantically guided frontier-selection policy designed for indoor environments, which prioritizes narrow openings and doors as informative frontiers to transition between semantically different regions while directing exploration toward relevant visual targets and goals.
  \item To ensure efficient scaling to multi-target questions, we introduce a bounded memory that selectively retains target-specific visual snapshots, enabling lightweight operation.
\end{itemize}

\section{Related Works}
\label{sec:related_works}

\begin{figure*}[!t]
    \centering
    \includegraphics[width=\textwidth]{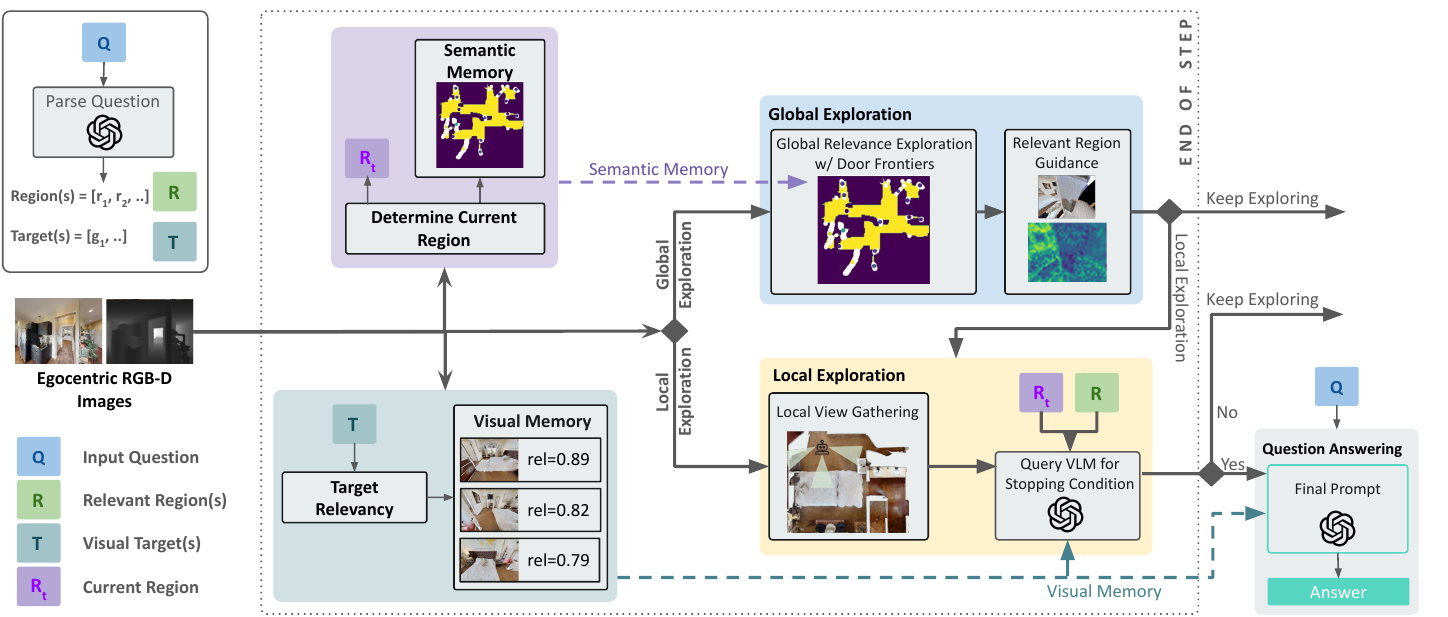}
    \caption{FAST-EQA processes the question (Q) by extracting relevant regions (R) and visual targets (T). At each step, it localizes its current region (\( \mathrm{R}_{\mathrm{t}} \)) and updates a semantic memory. For each target $T_m$, it maintains a dedicated memory $\psi_m$ that is refined using a visual relevance score. The agent employs a semantic frontier–guided global exploration strategy, leveraging narrow passages (e.g., doors and hallways) to effectively search for relevant semantic regions. When a relevant region is reached, it switches to local exploration to refine the target-specific memory. Once the stopping criterion is satisfied, the agent queries a VLM to generate the final answer. Dotted lines indicate module inputs while solid lines indicate procedural direction of the system.}
    \label{fig:system}
\end{figure*}

\subsection{Embodied Question Answering}
Embodied Question Answering (EQA) tasks agents with navigating environments to answer language queries. Early works such as EmbodiedQA \cite{EQA} and IQA \cite{IQA} established the problem of coupling active perception with language grounding. Hierarchical methods like Neural Modular Control \cite{EQA2} decompose navigation into semantic subgoals, while Multi-Target EQA\cite{MTEQA} extended reasoning to comparative queries across multiple objects. Subsequent advances incorporated richer perception: point cloud inputs improved photorealistic navigation \cite{EQA3}, and VideoNavQA \cite{VideoNavQA} highlighted reasoning challenges even when navigation is removed. Recent efforts emphasize efficiency and generalization. EfficientEQA \cite{EfficientEQA} and HM-EQA \cite{HMEQA} introduced semantic-guided exploration and early-stopping, while Fine-EQA \cite{FineEQA} benchmarked fine grained exploration-aware answering. Foundation model integration has also emerged: OpenEQA reframed EQA in the context of VLMs, and graph- or memory-augmented approaches (e.g., GraphEQA \cite{GraphEQA}, 3D-Mem \cite{3DMem}, MemoryEQA \cite{MemoryEQA}) addresses long-horizon reasoning. Collectively, these works show progress toward real-world EQA but reveal persistent gaps in spatial reasoning, structured memory for complex queries, and exploration time.

\subsection{VLMs for Embodied Tasks}
Large-scale vision-language models (VLMs) have been adapted to embodied contexts to improve reasoning. PaLM-E \cite{PALME} demonstrated that grounding multimodal LLMs in sensory inputs enables general-purpose planning and perception, with transfer across VQA and robotics. SpatialVLM \cite{SpatialVLM} addressed VLMs’ weakness in 3D reasoning by training on large-scale spatial QA, improving distance and relation understanding. Complementary to this, Think, Act, and Ask \cite{TAA} showed how LLMs can orchestrate exploration and dialog for personalized navigation, highlighting the potential of interactive reasoning in embodied tasks. While these works showcase the versatility of foundation models, their effectiveness often depends on coupling with structured spatial memory for reliable reasoning.

\subsection{Memory Representations for Embodied Tasks}
Effective memory remains central to Embodied QA. Metric- and semantic-based approaches like Semantic MapNet \cite{SemanticMapNet} and Goal-Oriented Semantic Exploration \cite{GoalOriented} build allocentric maps that accumulate semantic cues for navigation and object search. Semi-Parametric Topological Memory and VLMaps further demonstrated that topological and vision-language fused maps support robust localization and zero-shot navigation. Extensions such as VLFM \cite{VLFM} leverage language-grounded frontier exploration, while ConceptGraphs  \cite{ConceptGraphs} and GraphEQA \cite{GraphEQA} use 3D semantic scene graphs for relational reasoning. Memory-centric designs, e.g. MemoryEQA \cite{MemoryEQA} and ReMEmbR \cite{Remembr}, place structured, queryable memory at the core of decision-making, enabling long-horizon and temporal reasoning. These advances confirm that semantically enriched memory is key to bridging perception and reasoning, but also underscore challenges in scaling memory to open-vocabulary, complex environments.

\section{Problem Formulation}
\label{sec:problem_formulation}

We formulate the problem of EQA similar to \cite{HMEQA, MemoryEQA}. For an EQA task, an agent is initialized at a random location within a given single-floor environment and asked a natural language question. In any given episode, the agent will explore up to \(t \leq N\) timesteps where \(N\) is the maximum time steps allowed for the robot, which is scaled relative to the size of the scene. Each action step corresponds to a discrete change in the agent’s pose, either through translation or rotation, and planning of the next pose. 
An EQA scenario then consists of a sequence of RGB-D observations \(O = \{o_i\}_{i=0}^N\) obtained from exploration, a state sequence of agent poses during exploration, \(S = \{s_i\}_{i=0}^N\), an input question \(Q\), and some ground-truth answer \(y\). In this work, we consider two types of questions: multiple-choice, where the set of candidate answers is guaranteed to contain the correct label \(y\), and open-ended, where the answer is expressed as a free-form text sentence.


Furthermore, in our method we define the relevant room(s) for exploration for a given query as a set \(R\). The visual target(s) that needs to be observed are denoted as \(T = \{T_m\}_{m=0}^M\) where \(M\) is the total number of visual targets, and the current region the agent sees at timestep \(t\) is denoted as \(R_t\). For example, for the query \textit{``Is the bedsheet color in the master bedroom the same as in the guest bedroom''}, the relevant regions would be \( R = \{\text{\texttt{'master bedroom'}}, \text{\texttt{'guest bedroom'}}\} \) and the visual targets would be \(T = \{\texttt{bedsheet}_{\texttt{master}}, \texttt{bedsheet}_{\texttt{guest}}\}\). 
The scenario concludes upon satisfaction of the stopping criterion, which, in our setup, corresponds to prompting GPT-4o with stopping prompt and the top-$k$ relevant images. All the prompts are described in the Supplementary Material.
\section{Methodology}
\label{sec:methodology}
The FAST-EQA framework consists of global and local relevancy-guided exploration, visual memory retrieval, and chain-of-thought reasoning. The system diagram for how the framework operates in a given timestep $t$, is shown in Figure \ref{fig:system}. 

At the core of EQA methods is the exploration policy used. Existing methods for EQA have often relied on exploration policies that ask a VLM to select the next best direction from an annotated set of points on the current 2D observation to semantically weight frontiers \cite{HMEQA, MemoryEQA}. In most cases however, scenes have multiple rooms and the agent is often not be in a region relevant to the question. Thus, selecting the next direction in this way can be both inefficient and unintuitive. Given a certain query, humans already have some prior belief of which regions may be relevant and can quickly explore each of these regions. E.g. if asked about a bed in an unknown scene, the most natural strategy is to look for bedrooms, eliminating rooms that are not a bedroom and going through hallways to reach where a bedroom may be. Inspired by this, our method first leverages an LLM to parse the input query and select which room types may be relevant for exploration. For explicit queries such as \textit{``Is the kitchen tap on?''}, the relevant room would be \texttt{\{'kitchen'\}}. For implicit queries such as \textit{``Where did I leave my blanket?''}, the LLM is asked to guess which rooms may be relevant, returning \texttt{\{'bedroom', 'living room'\}}. 
This extracted information acts as an information filter for the subsequent exploration, narrowing the search space and focusing the exploration on observing the correct global region(s) first. From it's initial location, we also spin the agent 360 degrees to capture a panoramic view of the scene and form an initial occupancy map. Each subsequent exploration step follows either Global Relevance Exploration or Local Relevance Exploration, which are treated as distinct step types.


\subsection{Global Relevance Exploration}

For agent's global exploration, we leverage a lightweight 3D voxel representation with fixed height only for occupancy mapping and exploration tracking. Voxels are updated in real-time through Truncated Signed Distance Function (TSDF) fusion \cite{zeng20163dmatch} as the agent explores, tracking both occupied voxels and explored voxels within view using the input depth. Leveraging the occupancy map obtained, we implement a detector that identifies navigable narrow openings in the scene, which likely indicate doorways or hallways that are informative transitions between regions that are semantically different. To determine narrow openings that are likely doorways and hallways, we take a 2D slice of the 3D voxelized occupancy grid at the height of the agent and use the scene structure to determine candidate voxels \(\mathcal{V}\) by checking the occupancy of each voxel in the slice and the voxels within $\Delta x$, and $\Delta y$ directions nearby. Here \(\phi(v)\) is the occupancy of voxel \(v\). 


\begin{equation}
\mathcal{V} = \Big\{ v \;\Big|\; 
(\phi(v) = 0 \;\wedge\;
\phi(v \pm \Delta x) = 1
\;\lor\;
\phi(v \pm \Delta y) = 1)
\Big\}
\end{equation}

We then cluster these candidate voxels using the DBSCAN algorithm \cite{ester1996density} and use the cluster centroids as our global exploration frontiers, \(\mathcal{F} = \{f_1, f_2, ...,f_J\}\) for varying number of frontiers, \(J\). We maintain these frontiers in a priority queue, prioritizing frontiers that are a) formed from larger clusters and b) closer to unexplored regions. With this global exploration method, our agent is able to efficiently transition between semantically different regions in an environment to maximize coverage of the scene and the likelihood of traveling to question-relevant regions. 

\begin{figure}[t]
    \raggedright
    \includegraphics[width=\linewidth]{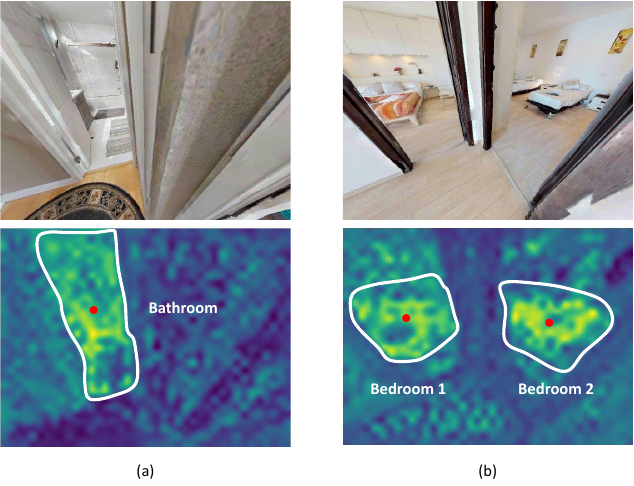}
    \caption{FAST-EQA leverages language-aligned features from AM-RADIO together with a SigLIP adaptor, to direct the agent toward the target regions. For queries such as (a) Bathroom and (b) Bedroom, the predicted heatmaps are thresholded to produce white contour segments, while the red dot indicates the contour centroid to step towards. This visualization illustrates how semantic grounding enables precise localization of task-relevant areas in the environment to guide exploration from global to local.}
    \label{fig:door_feature}
\end{figure}

At every step of exploration, we query a VLM with the current observation \(o_t\) to output what room(s) are seen from the agent's current perspective. For the VLM, we leverage Prismatic-VLM \cite{karamcheti2024prismatic}, which is a 7B parameter model that can be queried locally, to account for the high computational cost of querying this VLM at every step.
If a relevant room is detected from the agent's current position but the agent is outside the room, we leverage language-aligned feature guidance to first guide the agent inside the relevant room. As shown in Figure \ref{fig:door_feature}, the 2D image is featurized using AM-RADIO with a SigLIP adaptor \cite{ranzinger2024radio} and queried with the relevant room name  to obtain language-aligned features to obtain a contour segment for the target room to move into. The centroid of this contour is then projected into 3D space using depth information, producing a direction vector that points the agent toward the room, enabling efficient and semantically grounded entry.

\subsection{Local Relevance Exploration}

Once the agent is inside a relevant room or region, the primary goal of the agent is to gather local views in search of the visual target. We spin the agent around 360 degrees to gather a panoramic viewpoint of the room. The stop condition is queried at each timestep \(t\) with the visual memory and current observation if the agent is still facing towards the region, determined by checking that the detected region at the current timestep \(t\) is from the set of relevant regions, \(R_t \in R\). Each time the stop condition is queried, the agent can choose whether to stop exploration—if it can already answer the question—or to continue exploring. If the agent chooses to stop, relevant visual memory is retrieved for question-answering as detailed below. We use GPT-4o \cite{hurst2024gpt} as the VLM. 


\subsection{Relevant Memory Retrieval}
\begin{figure}[t]
    \raggedleft
    \includegraphics[width=\linewidth]{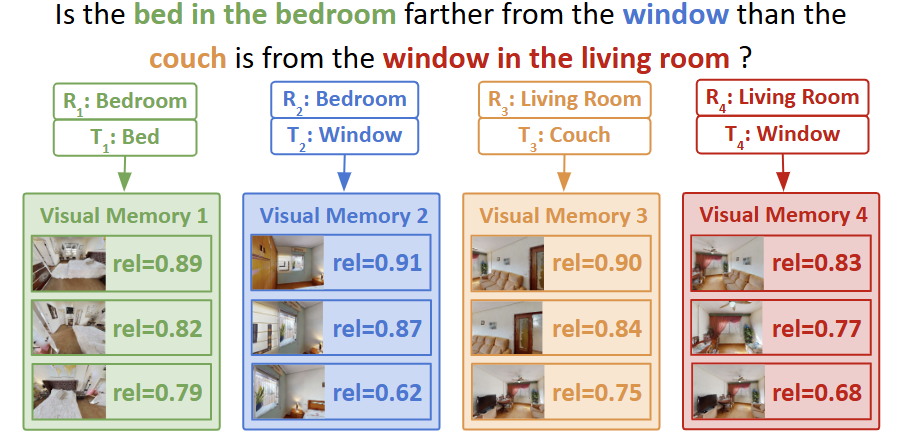}
    \caption{FAST-EQA employs a bounded memory system that allocates a dedicated visual memory for each target, retaining the $k$ most relevant images (here, $k = 3$). The overall memory footprint scales only with the number of targets and remains constant over time, even in long-horizon tasks.}
    \label{fig:memory}
\end{figure}

Throughout exploration, we maintain a lightweight visual memory, \(\psi\), that selectively retains the $k$ most relevant visual observations seen so far based on the visual relevance score, \(rel_{vis}\) of the observation (\ref{fig:memory}). This memory is stored per visual target, \(T_m\).
\begin{equation}
\psi_{T_m} = TopK_{o_t \in O}(rel_{vis}(o_t, Q, T_m))
\end{equation}
The selective retention of only the most relevant visual observations results in a bounded memory that scales directly to the number of targets $M$, extracted from the question. Upon either querying of the stop condition or final question-answering, the stored visual observations are retrieved to assist the VLM in making a decision. To rank the relevance of each observation, we compute the visual relevance score \(rel_{vis}(o_t, Q, T_m)\), by a weighted combination of a relevance score obtained from the CLIP \cite{radford2021learning} embedding model \(rel_{CLIP}(o_t, T_m)\), and the relevance score given by querying a generative VLM \cite{karamcheti2024prismatic} \(rel_{VLM}(o_t, Q)\), where the weight is denoted \(\lambda\).


\begin{equation}
\begin{split}
rel_{vis}(o_t, Q, T_m) &= \lambda \cdot rel_{CLIP}(o_t, T_m) \\
&\quad + (1 - \lambda) \cdot rel_{VLM}(o_t, Q)
\end{split}
\label{eq:3}
\end{equation}

The CLIP relevance score \(rel_{CLIP}\), calculated per visual target, is obtained by computing the cosine embedding similarity between the 2D RGB observation \(o_t\) and the text of the target object, \(T_m\) extracted from the input question, where \(f\) is the CLIP encoder.
\begin{equation}
rel_{CLIP} (o_t, T_m) = sim(o_t, T_m) = \frac{ f_{\text{text}}(T_m) \cdot f_{\text{img}}(o_t) }
{ \lVert f_{\text{text}}(T_m) \rVert \; \lVert f_{\text{img}}(o_t) \rVert}
\label{eq:4}
\end{equation}


The relevancy score from a generative VLM, \(rel_{VLM}\), is question-dependent and computed by prompting the model \(\hat{f}\) with the input question $Q$ and the agent’s current observation \(o_{t}\). Following \cite{HMEQA}, we use a prompt \(P\) that asks the model to judge whether the observation contains sufficient evidence to answer the question, and we take the probability assigned to the response ``yes'' as the relevancy score. For \(\hat{f}\), we employ Prismatic-VLM as our generative backbone.

\begin{equation}
rel_{VLM}(o_t, Q) = p(``yes" | \hat{f}(P(Q), o_t))
\end{equation}
Overall, the CLIP relevance score can be viewed as a focused quantification of whether the observation is related to the visual target, while the VLM relevance score is a holistic scoring of the observation given the entire context of the question. This combination scoring allows the agent to retrieve observations that align with both the focused target goal and the question-answering goal. We tune the weight \(\lambda\) through a hyperparameter search (Supplementary Sec. 3).

\subsection{Chain-of-Thought Question Answering}
In recent years, Chain-of-Thought (CoT) reasoning \cite{CoT} has proven effective in natural language and vision–language tasks \cite{MM_CoT}, but remains underexplored in embodied AI \cite{Monologue}. Multi-target question scenarios are one example that require reasoning over diverse observations, which naturally aligns with the compositional reasoning abilities of CoT. This motivates the use of CoT prompting for the final QA module.

\begin{figure}[t]
    \raggedleft
    \includegraphics[width=\linewidth]{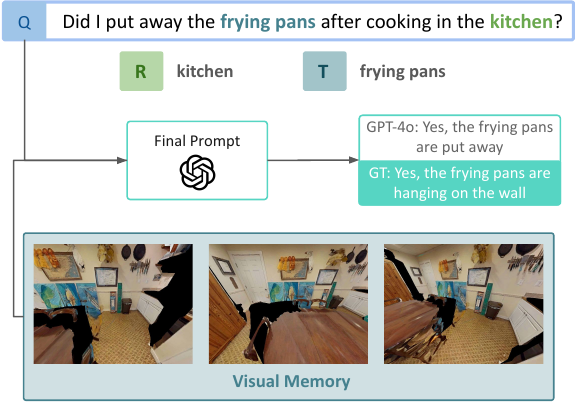}
    \caption{An example from EXPRESS-Bench illustrating how FAST-EQA identifies the relevant region $R$ and target $T$ from the question $Q$. It then explores the scene, and once the stopping condition is met, correctly generates the final answer from the retrieved visual memory.}
    \label{fig:final}
\end{figure}

Given the input question \(Q\) and the set of $M \times k$ observations retained in memory, we prompt a multi-frame VLM at the end of exploration to generate an answer either as a letter for multiple-choice QA, or an open-ended text response (Figure \ref{fig:final}) and asking the VLM to explicitly think step by step. This not only improves reasoning quality but also leads to more interpretable answers referring explicitly to the given visual observations, which existing baselines often lack. For the answering model, we use GPT-4o.

\begin{table*}[!t]
\centering
\caption{Performance comparison across EQA benchmarks against SOTA baseline methods. * indicates that the result is from a reproduced experiment reported by others. $^\dagger$ indicates results are on full A-EQA split.}
\label{tab:results}
\small
\resizebox{\textwidth}{!}{
\begin{tabularx}{\textwidth}{p{2.8cm} *{8}{>{\centering\arraybackslash}m{1.4cm}}}
\toprule
 & \multicolumn{2}{c}{HM-EQA} & \multicolumn{2}{c}{MT-HM3D} & \multicolumn{2}{c}{EXPRESS-Bench} & \multicolumn{2}{c}{A-EQA (184)} \\
\cmidrule(lr){2-3} \cmidrule(lr){4-5} \cmidrule(lr){6-7} \cmidrule(lr){8-9}
 & SR & Steps ($\downarrow$) 
 & SR & Steps ($\downarrow$)
 & LLM Score & $E_{path}$ ($\uparrow$)
 & LLM-Match & $E_{path}$ ($\uparrow$) \\
\midrule
GPT-4V (OpenEQA)\cite{OpenEQA} & -   & -   & -     & -   & -    & -      & 41.8 & 7.5 \\
Explore-EQA \cite{HMEQA} & 58.4 & 0.52 & 36.2*  & 0.64 & -    & -      & 46.9* & 23.4 \\
Graph-EQA \cite{GraphEQA} & 63.5 & \textbf{0.20} & 45.63* & 0.45 & -    & -      & 30.1*{$^\dagger$}  & - \\
Memory-EQA \cite{MemoryEQA} & 63.4  & 0.40  & \textbf{55.1}   & \textbf{0.41} & -    & -      & 36.8{$^\dagger$}  & - \\ 
Fine-EQA \cite{FineEQA} & 56.0 & 0.54 & -     & -   & 63.95 & 25.58 & 43.3{$^\dagger$} & 29.2 \\
3D-Mem \cite{3DMem} & 50.40   & 0.63   & -     & -   & -    & -      & \textbf{52.6} & \textbf{42.0} \\
\midrule
FAST-EQA (Ours) & \makecell{$\mathbf{69.2}$ \\ $\pm 0.7$} 
         & \makecell{$0.65$ \\ $\pm 0.01$} 
         & \makecell{$50.5$ \\ $\pm 0.3$} 
         & \makecell{$0.52$ \\ $\pm 0.01$} 
         & \makecell{$\mathbf{68.7}$ \\ $\pm 0.5$} 
         & \makecell{$\mathbf{29.25}$ \\ $\pm 0.55$} 
         & \makecell{$49.0$ \\ $\pm 1.7$} 
         & \makecell{$27.70$ \\ $\pm 1.70$} \\
\bottomrule
\end{tabularx}
}
\end{table*}

\section{Evaluation and Results}
\label{sec:eval}

We conduct experiments across diverse EQA benchmarks, keeping the definition of an exploration step consistent. Following the parameter set in \cite{HMEQA}, the agent is allowed to travel up to a maximum of 3 meters per translational step.

\subsection{Comparison Across Benchmarks}
To evaluate our approach, we run FAST-EQA on four benchmarks: HM-EQA \cite{HMEQA}, MT-HM3D \cite{MTEQA}, EXPRESS-Bench \cite{FineEQA}, and the released A-EQA 184-split of OpenEQA \cite{OpenEQA}, which covers diverse question formats and question types. HM-EQA and MT-HM3D contain multiple-choice questions, with MT-HM3D focusing on questions comparing multiple targets. EXPRESS-Bench and OpenEQA consist of open-ended questions, with EXPRESS-Bench being the largest scale benchmark, containing over 2000 questions. To the best of our knowledge, we conduct the most comprehensive evaluation on diverse benchmarks compared to existing methods. Furthermore, we run three trials on each dataset, with temperature $\tau=0$, to obtain error bounds for our results, which takes into consideration any stochasticity in our method. We note that existing methods do not report error bounds, despite the usage of VLMs and LLMs often resulting in such variance.

For the multiple-choice datasets we report success rate (SR) and normalized steps (Steps) following \cite{HMEQA, MemoryEQA}. For the open-vocabulary questions, we report the LLM-Score following \cite{FineEQA} for EXPRESS-Bench, LLM-Match (\(C\)) following \cite{OpenEQA} for OpenEQA, and the corresponding path efficiency metric calculated with success weighted-by-path-length (SPL) using geodesic distance, \(E_{path}\). For more details on metrics, refer to supplementary material.

\spar{Results.} The performance of FAST-EQA against a number of SOTA baselines is shown in Table \ref{tab:results}. We observe that our method surpasses existing methods on question-answering accuracy for both HM-EQA and EXPRESS-Bench, highlighting its ability to handle both multiple-choice and open-ended questions. In particular, we achieve a 9\% improvement over the next-best method on HM-EQA and a 7\% improvement over the SOTA method on EXPRESS-Bench, marking high performance on a larger-scale dataset. On the A-EQA subset of Open-EQA, we achieve results comparable to SOTA. While multi-target questions still pose a challenge, FAST-EQA is also able to achieve competitive performance against existing methods with relatively low variance while having a lightweight, bounded memory.

We further note that our method prioritizes question-answering accuracy against the step length tradeoff, allowing for more thorough exploration and gathering of diverse views before generating an answer. Simultaneously, the inference time speed-up we achieve maintains overall efficiency.

\subsection{Inference Time Experiments}

For EQA systems, the ability to operate in real-time is essential for eventual deployment on physical robots. Existing EQA methods however, do not report these evaluations. To address this gap, we benchmark and compare the inference time of our method with existing open-source methods. Inference time experiments are run for three trials on a fixed subset of 100 questions from the HM-EQA dataset. All experiments are run on a single NVIDIA H100 Tensor Core GPU with 80GB RAM. We report the average time (in seconds) for each exploration step, which includes the time taken for querying large models. 

Table \ref{tab:inf-time} shows the inference time results. Our method achieves significantly faster inference times than existing EQA methods, resulting in a 13.6\% speedup over the next fastest method. Furthermore, FAST-EQA achieves consistently fast run times, exhibiting markedly lower variance across trials than competing methods.

\begin{table}
    \centering
    \begin{tabular}{lc}
    \toprule
         & Average Step Time (s) \\
        \midrule
        Fine-EQA  &  $2.94 \pm 0.14$\\
       Explore-EQA  &  $4.90 \pm 0.26$\\
       3D-Mem  &  $14.53 \pm 0.25$\\
       Memory-EQA  &  $15.89 \pm 0.65$\\
       \midrule
       FAST-EQA (Ours) & $\mathbf{2.54 \pm 0.05}$\\
    \bottomrule
    \end{tabular}
    \caption{Average step time, in seconds, of FAST-EQA compared to other methods on a subset of questions from HM-EQA}
    \label{tab:inf-time}
\end{table}

\subsection{Ablation Studies}
We further conduct a series of ablation studies to isolate the contributions of individual  design choices in our framework. We run these experiments on a fixed sampled subset of 100 questions from HM-EQA, and summarize the results in Table \ref{tab:ablation}. Each ablation removes one design choice at a time and compares against the full FAST-EQA model.

\spar{Without Doorways/Openings as Frontiers.} {To evaluate the role of structural scene analysis in exploration, we replace our doorway- and opening-guided strategy with standard frontier-based exploration (FBE) \cite{FBE}, where frontiers are defined purely as boundaries between explored and unexplored space.}

\spar{Without VLM Relevancy.} To quantify the effect of incorporating VLM confidence into scoring observation relevancy, we experiment with scoring the relevance of image observations solely based on the CLIP similarity to the visual target by setting $\lambda=1$ in Equation \ref{eq:3}.

\spar{Without CLIP Relevancy.} Similarly, we ablate the inclusion of CLIP similarity scores when scoring observation relevancy. We score the relevance of image observations only based on querying the VLM for whether the image is relevant by setting $\lambda=0$ in Equation \ref{eq:3}.

\spar{Without Chain-of-Thought Prompting.} 
We ablate the effect of using chain-of-thought prompting for the question-answering module by modifying the prompt to remove explicit instructions to ``think step by step''.

\spar{Results.} 
As shown in Table \ref{tab:ablation}, exclusion of doorway-guided frontier exploration, VLM relevancy scoring, CLIP scoring, or chain-of-thought prompting all result in a performance drop for overall question-answering success, highlighting the importance of each component. Furthermore the semantic frontier-guided exploration leveraging doorways and openings has the largest effect for improving the performance, as changing to standard FBE results in a 9\% drop in success rate. This demonstrates that semantic exploration guided by informative structural information is particularly critical for enabling efficient and accurate embodied QA. More broadly, these ablations validate our main contribution: combining semantics-aware exploration together with multimodal relevancy scoring and reasoning mechanisms yields substantial gains for accurate question-answering without compromising computation time.

\begin{table}
    \centering
    \begin{tabular}{lcc}
    \toprule
        & SR & Steps ($\downarrow$)\\
    \midrule
       FAST-EQA w/o Doorway Frontiers & 67.0 & 0.64 \\
       FAST-EQA w/o CLIP Scoring  & 69.0 & 0.63 \\
       FAST-EQA w/o VLM Scoring  & 72.0 & 0.63 \\
       FAST-EQA w/o CoT Reasoning & 72.0 & 0.60 \\
       \midrule
       FAST-EQA (Ours)  & \textbf{76.0} & 0.65 \\
    \bottomrule
    \end{tabular}
    \caption{Ablation of various components of our EQA system on a fixed subset of HM-EQA}
    \label{tab:ablation}
\end{table}

\section{Conclusion}
\label{sec:conclusion}

We introduce FAST-EQA, an active exploration QA framework that couples semantically-guided global and local exploration
with a bounded memory that scales only with targets, making it well-suited for single as well as multi-target tasks. Our exploration strategy is focused on intuitively maximizing exploration coverage by leveraging scene morphology to transition between semantic regions, allowing for thorough search and verification of targets. The framework supports both multiple-choice and open-ended question settings, as well as single- and multi-target questions. Across major EQA benchmarks, FAST-EQA achieves state-of-the-art results on HMEQA and EXPRESS-Bench, while remaining competitive on OpenEQA and MT-HM3D. Coupled with this performance, our method delivers fast inference with low memory usage, enabling real-time deployment for embodied agents.

\subsection{Limitations and Future Work}

While FAST-EQA demonstrates strong performance across multiple EQA benchmarks, several limitations remain. First, the system’s effectiveness is bounded by the spatial reasoning capabilities of the current VLMs used. As highlighted by \cite{ThinkingInSpace}, even state-of-the-art MLLMs struggle with fine-grained spatial understanding, which in turn constrains the quality of reasoning in complex environments. Second, we observe that the VLM (e.g., GPT-4o) used for reasoning exhibits variance across runs, sometimes producing inconsistent answers. In certain cases, this variability can cause the agent to terminate prematurely, limiting robustness and reliability in long-horizon tasks.

Looking ahead, an exciting direction is the development of latent visual memory representations that can efficiently capture and compress both short-term and long-term scene information. Such memory mechanisms could enable more consistent reasoning, reduce reliance on repetitive VLM queries, and improve the agent’s ability to handle extended exploration tasks. We believe integrating structured spatial memory with language-driven reasoning offers a promising pathway toward more generalizable and reliable embodied agents.
{
    \small
    \bibliographystyle{ieeenat_fullname}
    \bibliography{main}

@String(CVPR= {IEEE Conf. Comput. Vis. Pattern Recog.})

@String(BMVC= {Brit. Mach. Vis. Conf.})

@String(AAAI = {AAAI})

@String(CVPR  = {CVPR})

@String(BMVC  =	{BMVC})

@inproceedings{EQA,
  author    = {Abhishek Das and Samyak Datta and Georgia Gkioxari and Stefan Lee and Devi Parikh and Dhruv Batra},
  title     = {Embodied Question Answering},
  booktitle = {Proceedings of the IEEE Conference on Computer Vision and Pattern Recognition (CVPR)},
  year      = {2018},
  month     = jun,
  pages     = {1--10},
  address   = {Salt Lake City, UT, USA}
}

@inproceedings{IQA,
  author      = "Daniel Gordon and Aniruddha Kembhavi and Mohammad Rastegari and Joseph Redmon and Dieter Fox and Ali Farhadi",
  title       = "IQA: Visual Question Answering in Interactive Environments",
  booktitle   = "Proceedings of the IEEE Conference on Computer Vision and Pattern Recognition (CVPR)",
  address     = "Salt Lake City, UT, USA",
  month       = jun,
  year        = "2018",
  pages       = "4089--4098"
}

@inproceedings{EQA2,
  author    = "Abhishek Das and Georgia Gkioxari and Stefan Lee and Devi Parikh and Dhruv Batra",
  title     = "Neural Modular Control for Embodied Question Answering",
  booktitle = "Proceedings of The 2nd Conference on Robot Learning (CoRL)",
  address   = "Zürich, Switzerland",
  month     = oct,
  year      = "2018",
  pages     = "53--62"
}

@inproceedings{EQA3,
  author      = "Erik Wijmans and Samyak Datta and Oleksandr Maksymets and Abhishek Das and Georgia Gkioxari and Stefan Lee and Irfan Essa and Devi Parikh and Dhruv Batra",
  title       = "Embodied Question Answering in Photorealistic Environments with Point Cloud Perception",
  booktitle   = "Proceedings of the IEEE/CVF Conference on Computer Vision and Pattern Recognition (CVPR)",
  address     = "Long Beach, CA, USA",
  month       = jun,
  year        = "2019",
  pages       = "6659--6668"
}

@inproceedings{MTEQA,
  author    = {Licheng Yu and Xinlei Chen and Georgia Gkioxari and Mohit Bansal and Tamara L. Berg and Dhruv Batra},
  title     = {Multi-Target Embodied Question Answering},
  booktitle = {Proceedings of the IEEE/CVF Conference on Computer Vision and Pattern Recognition (CVPR)},
  year      = {2019},
  month     = {June},
  pages     = {4990--4999},
  doi       = {10.1109/CVPR.2019.00511},
  url       = {https://arxiv.org/abs/1904.04686}
}

@inproceedings{PALME,
  author      = "Danny Driess and Fei Xia and Mehdi S. M. Sajjadi and Corey Lynch and Aakanksha Chowdhery and Brian Ichter and Ayzaan Wahid and Jonathan Tompson and Quan Vuong and Tianhe Yu and Wenlong Huang and Yevgen Chebotar and Pierre Sermanet and Daniel Duckworth and Sergey Levine and Vincent Vanhoucke and Karol Hausman and Marc Toussaint and Klaus Greff and Andy Zeng and Igor Mordatch and Pete Florence",
  title       = "PaLM-E: An Embodied Multimodal Language Model",
  booktitle   = "Proceedings of the 40th International Conference on Machine Learning (ICML)",
  address     = "Honolulu, HI, USA",
  month       = jul,
  year        = "2023",
  pages       = "8469--8488"
}

@inproceedings{SpatialVLM,
  author      = "Boyuan Chen and Zhuo Xu and Sean Kirmani and Brain Ichter and Dorsa Sadigh and Leonidas Guibas and Fei Xia",
  title       = "SpatialVLM: Endowing Vision-Language Models with Spatial Reasoning Capabilities",
  booktitle   = "Proceedings of the IEEE/CVF Conference on Computer Vision and Pattern Recognition (CVPR)",
  address     = "Vancouver, BC, Canada",
  month       = jun,
  year        = "2024",
  pages       = "14455--14465"
}

@inproceedings{TAA,
  author      = "Yinpei Dai and Run Peng and Sikai Li and Joyce Chai",
  title       = "Think, Act, and Ask: Open-World Interactive Personalized Robot Navigation",
  booktitle   = "Proceedings of the IEEE International Conference on Robotics and Automation (ICRA)",
  address     = "Singapore",
  month       = may,
  year        = "2024",
  pages       = "TBD"
}

@inproceedings{Remembr,
  author    = {Abrar Anwar and John Welsh and Joydeep Biswas and Soha Pouya and Yan Chang},
  title     = {ReMEmbR: Building and Reasoning Over Long‑Horizon Spatio‑Temporal Memory for Robot Navigation},
  booktitle = {arXiv preprint arXiv:2409.13682},
  year      = {2024},
  month     = sep
}

@inproceedings{VideoNavQA,
  author    = {Catalina Cangea and Eugene Belilovsky and Pietro Lio and Aaron Courville},
  title     = {VideoNavQA: Bridging the Gap between Visual and Embodied Question Answering},
  booktitle = {Proceedings of the 30th British Machine Vision Conference (BMVC)},
  year      = {2019},
  note      = {Spotlight presentation at Visually Grounded Interaction and Learning (ViGIL) Workshop, NeurIPS 2019},
  doi       = {10.17863/CAM.44469},
  url       = {https://arxiv.org/abs/1908.04950}
}

@inproceedings{SemanticMapNet,
  author    = {Vincent Cartillier and Zhile Ren and Neha Jain and Stefan Lee and Irfan Essa and Dhruv Batra},
  title     = {Semantic MapNet: Building Allocentric Semantic Maps and Representations from Egocentric Views},
  booktitle = {Proceedings of the AAAI Conference on Artificial Intelligence},
  year      = {2021},
  volume    = {35},
  number    = {2},
  pages     = {964--972},
  doi       = {10.1609/aaai.v35i2.16180},
  month     = may
}

@inproceedings{GoalOriented,
  author    = {Devendra Singh Chaplot and Dhiraj Gandhi and Abhinav Gupta and Ruslan Salakhutdinov},
  title     = {Object Goal Navigation using Goal-Oriented Semantic Exploration},
  booktitle = {Advances in Neural Information Processing Systems (NeurIPS)},
  year      = {2020},
  url       = {https://arxiv.org/abs/2007.00643},
  note      = {Winner of the CVPR 2020 Habitat ObjectNav Challenge},
  month     = {December}
}

@inproceedings{EfficientEQA,
  author    = {Kai Cheng and Zhengyuan Li and Xingpeng Sun and Byung-Cheol Min and Amrit Singh Bedi and Aniket Bera},
  title     = {EfficientEQA: An Efficient Approach for Open Vocabulary Embodied Question Answering},
  booktitle = {Proceedings of the IEEE/CVF International Conference on Robotics and Automation (ICRA)},
  year      = {2025},
  month     = may,
  url       = {https://arxiv.org/abs/2410.20263},
  note      = {Accepted for publication},
  doi       = {10.1109/ICRA48506.2025.00001}
}

@inproceedings{OpenEQA,
  author    = {Arjun Majumdar and Anurag Ajay and Xiaohan Zhang and Pranav Putta and Sriram Yenamandra and Mikael Henaff and Sneha Silwal and Paul McVay and Oleksandr Maksymets and Sergio Arnaud and Karmesh Yadav and Qiyang Li and Ben Newman and Mohit Sharma and Vincent‑Pierre Berges and Shiqi Zhang and Pulkit Agrawal and Dhruv Batra and Yonatan Bisk and Mrinal Kalakrishnan and Franziska Meier and Chris Paxton and Alexander Sax and Aravind Rajeswaran},
  title     = {OpenEQA: Embodied Question Answering in the Era of Foundation Models},
  booktitle = {Proceedings of the IEEE/CVF Conference on Computer Vision and Pattern Recognition (CVPR)},
  year      = {2024}
}

@inproceedings{HMEQA,
  author    = {Allen Z. Ren and Jaden Clark and Anushri Dixit and Masha Itkina and Anirudha Majumdar and Dorsa Sadigh},
  title     = {Explore until Confident: Efficient Exploration for Embodied Question Answering},
  booktitle = {Proceedings of Robotics: Science and Systems},
  year      = {2024},
  month     = {July},
  address   = {Delft, Netherlands},
  doi       = {10.15607/RSS.2024.XX.089}
}

@inproceedings{FineEQA,
  author    = {Kaixuan Jiang and Yang Liu and Weixing Chen and Jingzhou Luo and Ziliang Chen and Ling Pan and Guanbin Li and Liang Lin},
  title     = {Beyond the Destination: A Novel Benchmark for Exploration‑Aware Embodied Question Answering},
  booktitle = {arXiv preprint arXiv:2503.11117},
  year      = {2025},
  month     = {March}
}

@inproceedings{3DMem,
  author    = {Yuncong Yang and Han Yang and Jiachen Zhou and Peihao Chen and Hongxin Zhang and Yilun Du and Chuang Gan},
  title     = {3D‑Mem: 3D Scene Memory for Embodied Exploration and Reasoning},
  booktitle = {Proceedings of the IEEE/CVF Conference on Computer Vision and Pattern Recognition (CVPR)},
  year      = {2025},
  month     = {June},
  pages     = {17294--17303}
}

@inproceedings{MemoryEQA,
  author    = {Mingliang Zhai and Zhi Gao and Yuwei Wu and Yunde Jia},
  title     = {Memory‑Centric Embodied Question Answer},
  booktitle = {arXiv preprint arXiv:2505.13948},
  year      = {2025},
  month     = may
}

@article{GraphEQA,
  author    = {Saxena, Saumya and Buchanan, Blake and Paxton, Chris and Chen, Bingqing and Vaskevicius, Narunas and Palmieri, Luigi and Francis, Jonathan and Kroemer, Oliver},
  title     = {GraphEQA: Using 3D Semantic Scene Graphs for Real-time Embodied Question Answering},
  journal   = {arXiv},
  year      = {2024},
  url       = {https://arxiv.org/abs/2412.14480},
  note      = {Accepted for publication},
  month     = {December}
}

@inproceedings{zeng20163dmatch, 
    title={3DMatch: Learning Local Geometric Descriptors from RGB-D Reconstructions}, 
    author={Zeng, Andy and Song, Shuran and Nie{\ss}ner, Matthias and Fisher, Matthew and Xiao, Jianxiong and Funkhouser, Thomas}, 
    booktitle={CVPR}, 
    year={2017} 
}

@inproceedings{karamcheti2024prismatic,
  title={Prismatic vlms: Investigating the design space of visually-conditioned language models},
  author={Karamcheti, Siddharth and Nair, Suraj and Balakrishna, Ashwin and Liang, Percy and Kollar, Thomas and Sadigh, Dorsa},
  booktitle={Forty-first International Conference on Machine Learning},
  year={2024}
}

@inproceedings{CLIP,
  author    = {Alec Radford and Jong Wook Kim and Chris Hallacy and Aditya Ramesh and Gabriel Goh and Sandhini Agarwal and Girish Sastry and Amanda Askell and Pamela Mishkin and Jack Clark and Gretchen Krueger and Ilya Sutskever},
  title     = {Learning Transferable Visual Models From Natural Language Supervision},
  booktitle = {Proceedings of the 38th International Conference on Machine Learning (ICML)},
  year      = {2021},
  pages     = {8748--8763},
  publisher = {PMLR},
  url       = {https://proceedings.mlr.press/v139/radford21a.html}
}

@inproceedings{FBE,
  author    = {Brian Yamauchi},
  title     = {A Frontier-Based Approach for Autonomous Exploration},
  booktitle = {Proceedings of the 1997 IEEE International Symposium on Computational Intelligence in Robotics and Automation (CIRA)},
  year      = {1997},
  pages     = {146--151},
  doi       = {10.1109/CIRA.1997.613851},
  url       = {https://doi.org/10.1109/CIRA.1997.613851},
  month     = {July},
  address   = {Monterey, CA, USA}
}

@article{hurst2024gpt,
  title={Gpt-4o system card},
  author={Hurst, Aaron and Lerer, Adam and Goucher, Adam P and Perelman, Adam and Ramesh, Aditya and Clark, Aidan and Ostrow, AJ and Welihinda, Akila and Hayes, Alan and Radford, Alec and others},
  journal={arXiv preprint arXiv:2410.21276},
  year={2024}
}

@inproceedings{ranzinger2024radio,
  title={Am-radio: Agglomerative vision foundation model reduce all domains into one},
  author={Ranzinger, Mike and Heinrich, Greg and Kautz, Jan and Molchanov, Pavlo},
  booktitle={Proceedings of the IEEE/CVF conference on computer vision and pattern recognition},
  pages={12490--12500},
  year={2024}
}

@inproceedings{radford2021learning,
  title={Learning transferable visual models from natural language supervision},
  author={Radford, Alec and Kim, Jong Wook and Hallacy, Chris and Ramesh, Aditya and Goh, Gabriel and Agarwal, Sandhini and Sastry, Girish and Askell, Amanda and Mishkin, Pamela and Clark, Jack and others},
  booktitle={International conference on machine learning},
  pages={8748--8763},
  year={2021},
  organization={PmLR}
}

@inproceedings{li2023blip,
  title={Blip-2: Bootstrapping language-image pre-training with frozen image encoders and large language models},
  author={Li, Junnan and Li, Dongxu and Savarese, Silvio and Hoi, Steven},
  booktitle={International conference on machine learning},
  pages={19730--19742},
  year={2023},
  organization={PMLR}
}

@article{liu2023visual,
  title={Visual instruction tuning},
  author={Liu, Haotian and Li, Chunyuan and Wu, Qingyang and Lee, Yong Jae},
  journal={Advances in neural information processing systems},
  volume={36},
  pages={34892--34916},
  year={2023}
}

@article{zhang2024navid,
  title={Navid: Video-based vlm plans the next step for vision-and-language navigation},
  author={Zhang, Jiazhao and Wang, Kunyu and Xu, Rongtao and Zhou, Gengze and Hong, Yicong and Fang, Xiaomeng and Wu, Qi and Zhang, Zhizheng and Wang, He},
  journal={arXiv preprint arXiv:2402.15852},
  year={2024}
}

@inproceedings{ester1996density,
  title={A density-based algorithm for discovering clusters in large spatial databases with noise},
  author={Ester, Martin and Kriegel, Hans-Peter and Sander, J{\"o}rg and Xu, Xiaowei and others},
  booktitle={kdd},
  volume={96},
  number={34},
  pages={226--231},
  year={1996}
}

@inproceedings{ConceptGraphs,
  author      = "Qiao Gu and Alihusein Kuwajerwala and Sacha Morin and Krishna Murthy Jatavallabhula and Bipasha Sen and Aditya Agarwal and Corban Rivera and William Paul and Kirsty Ellis and Rama Chellappa and Chuang Gan and Celso Miguel de Melo and Joshua B. Tenenbaum and Antonio Torralba and Florian Shkurti and Liam Paull",
  title       = "ConceptGraphs: Open-Vocabulary 3D Scene Graphs for Perception and Planning",
  booktitle   = "2024 IEEE International Conference on Robotics and Automation (ICRA)",
  address     = "Karlsruhe, Germany",
  month       = may,
  year        = "2024",
  pages       = "5021--5028"
}

@inproceedings{VLFM,
  author      = "Naoki Yokoyama and Sehoon Ha and Dhruv Batra and Jiuguang Wang and Bernadette Bucher",
  title       = "VLFM: Vision-Language Frontier Maps for Zero-Shot Semantic Navigation",
  booktitle   = "Proceedings of the IEEE International Conference on Robotics and Automation (ICRA)",
  address     = "Yokohama, Japan",
  month       = may,
  year        = "2024",
  pages       = "42--48"
}

@inproceedings{Hydra,
  author    = {Nathan Hughes and Yun Chang and Luca Carlone},
  title     = {Hydra: A Real‐time Spatial Perception System for 3D Scene Graph Construction and Optimization},
  booktitle = {Proceedings of Robotics: Science and Systems (RSS)},
  year      = {2022},
  month     = {June},
  address   = {New York City, NY, USA},
  doi       = {10.15607/RSS.2022.XVIII.050},
}

@inproceedings{HOVSG,
  author    = {Abdelrhman Werby and Chenguang Huang and Martin Büchner and Abhinav Valada and Wolfram Burgard},
  title     = {Hierarchical Open‐Vocabulary 3D Scene Graphs for Language-Grounded Robot Navigation},
  booktitle = {Proceedings of Robotics: Science and Systems},
  year      = {2024},
  address   = {Delft, Netherlands},
  month     = {July},
  doi       = {10.15607/RSS.2024.XX.077}
}

@article{OSG,
  title        = {Open Scene Graphs for Open-World Object-Goal Navigation},
  author       = {Joel Loo and Zhanxin Wu and David Hsu},
  journal      = {arXiv preprint arXiv:2508.04678},
  year         = {2025},
  month        = {August},
  note         = {Also to appear in *The International Journal of Robotics Research*},
  url          = {https://arxiv.org/abs/2508.04678}
}

@inproceedings{ThinkingInSpace,
  title        = {Thinking in Space: How Multimodal Large Language Models See, Remember, and Recall Spaces},
  author       = {Yang, Jihan and Yang, Shusheng and Gupta, Anjali W. and Han, Rilyn and Fei‐Fei, Li and Xie, Saining},
  booktitle    = {Proceedings of the IEEE/CVF Conference on Computer Vision and Pattern Recognition (CVPR)},
  year         = {2025},
  pages        = {10632--10643},
  doi          = {10.1109/CVPR64423.2025.01149},
  arxiv        = {2412.14171},
}

@inproceedings{CoT,
  author    = {Jason Wei and Xuezhi Wang and Dale Schuurmans and Maarten Bosma and Ed Chi and Quoc Le and Denny Zhou},
  title     = {Chain of Thought Prompting Elicits Reasoning in Large Language Models},
  booktitle = {Advances in Neural Information Processing Systems (NeurIPS)},
  year      = {2022},
  url       = {https://arxiv.org/abs/2201.11903}
}

@article{MM_CoT,
  author    = {Zhang, Zhuosheng and Xu, Aston Zhang and Li, Zhiyuan and Wang, Rui and Liu, Zhifang},
  title     = {Multimodal Chain-of-Thought Reasoning in Language Models},
  journal   = {arXiv preprint arXiv:2302.00923},
  year      = {2023},
  url       = {https://arxiv.org/abs/2302.00923}
}

@inproceedings{Monologue,
  author    = {Huang, Wenlong and Abbeel, Pieter and Pathak, Deepak and Mordatch, Igor},
  title     = {Inner Monologue: Embodied Reasoning through Planning with Language Models},
  booktitle = {Conference on Robot Learning (CoRL)},
  year      = {2022},
  url       = {https://arxiv.org/abs/2207.05608}
}

@inproceedings{wang2025embodied,
  title={Embodied scene understanding for vision language models via metavqa},
  author={Wang, Weizhen and Duan, Chenda and Peng, Zhenghao and Liu, Yuxin and Zhou, Bolei},
  booktitle={Proceedings of the Computer Vision and Pattern Recognition Conference},
  pages={22453--22464},
  year={2025}
}
}

\end{document}


\maketitle

\section{Evaluation Metrics}
The evaluation metrics used in the results Table \ref{tab:results} are calculated as follows, where \(N_{total}\) is the total number of questions:

\textbf{Success rate (SR)} for multiple-choice questions:
\[
SR = \frac{Correct}{N_{total}}\times 100\%
\]
where \(Correct\) is the number of questions answered correctly. For multiple-choice questions, we ask the model to output the letter corresponding to the choice.

\textbf{Normalized steps (Steps)} from MemoryEQA \cite{MemoryEQA}:
\[
Steps = \frac{1}{N_{total}} \sum_{i=1}^{N_{total}} \frac{q_i}{\sqrt{S_i * \gamma_s}} ,
\]
where \(q_i\) is the number of steps taken for question \(i\), \(S_i\) is the total room size, \(\gamma_s\) is the ratio between max steps and room size.

\textbf{LLM Score} from Fine-EQA \cite{FineEQA}:
\[
\text{\textit{LLM Score}} = \frac{1}{N_{total}} \sum_{i=1}^{N_{total}} \frac{\sigma_i}{5} \times 100\% .
\]
where \(\sigma_i\) is the raw score given by the LLM from 1 to 5. We use the same prompts as \cite{OpenEQA} for the LLM scoring procedure.

\textbf{LLM-Match} from OpenEQA \cite{OpenEQA}:
\[
\text{\textit{LLM-Match}} = \frac{1}{N_{total}} \sum_{i=1}^{N_{total}} \frac{\sigma_i - 1}{4} \times 100\% .
\]
where \(\sigma_i\) is the raw score given by the LLM from 1 to 5. We use the same prompts as OpenEQA \cite{OpenEQA} for the LLM scoring procedure.

\textbf{Path Efficiency ($E_{path}$)} (from OpenEQA \cite{OpenEQA} and Fine-EQA \cite{FineEQA}):
\[
E = \frac{1}{N_{total}} \sum_{i=1}^{N_{total}} \delta_i \times 
\frac{l_i}{\max(p_i, l_i)} \times 100\%,
\]
where \(\delta_i\) is the normalized score given by the LLM which is equal to \(\frac{\sigma_i - 1}{4}\) for OpenEQA and \(\frac{\sigma_i}{5}\) for Fine-EQA. \(l_i\) is the geodesic distance taken in the ground-truth trajectory and \(p_i\) is the distance traveled by the agent.

\section{Additional Ablation Results}

We conduct an ablation study on the size of our bounded visual memory, formed by the top-k relevant image observations per target. This visual memory is used for both determining the stopping condition and final question answering. We vary the value of parameter $k$ and measure the question-answering success rate and normalized steps taken in Table \ref{tab:top-k-full}. We observe that the success rate decreases when only one relevant image is retrieved, likely due to the lack of multi-view angles required for answering some questions. As we increase $k$, we see that performance increases, plateaus at $k=3$, and then achieves a peak at $k=5$. We evaluate FAST-EQA on all the datasets for both $k = 3$ as well as $5$. The results are reported in Table \ref{tab:results}, where we see that increasing the value of $k$ from 3 to 5 does not impact overall performance in any significant way. For HM-EQA and MT-HM3D, the success rate (SR) increases for $k=5$, whereas for EXPRESS-Bench, it stays the same. However, A-EQA presents contrary results. When we increase visual memory size, the LLM-Match score decreases. Since $k=3$ gives us the best overall results across all the benchmarks and reduces memory usage, we report results with $k=3$ in the main paper. 

\begin{table}
    \centering
    \begin{tabular}{lcc}
    \toprule
        & SR & Steps (↓)\\
    \midrule
       FAST-EQA$_{k=1}$ & 69.0 & 0.66 \\
       FAST-EQA$_{k=2}$ & 74.0 & 0.63 \\
       FAST-EQA$_{k=3}$ & 76.0 & 0.65 \\
       FAST-EQA$_{k=4}$  & 76.0 & 0.64 \\
       FAST-EQA$_{k=5}$  & 80.0 & 0.63 \\
       FAST-EQA$_{k=6}$  & 79.0 & 0.64 \\
    \bottomrule
    \end{tabular}
    \caption{Tuning the size $k$ of our bounded visual memory on a fixed subset of HM-EQA}
    \label{tab:top-k-full}
\end{table}

\begin{table*}[t]
\centering
\caption{Comparison across EQA benchmarks against SOTA baseline methods. * indicates that the result is from a reproduced experiment reported by others. $^\dagger$ indicates results are on full A-EQA split.}
\label{tab:results}
\small
\resizebox{\textwidth}{!}{
\begin{tabularx}{\textwidth}{p{2.8cm} *{8}{>{\centering\arraybackslash}m{1.4cm}}}
\toprule
 & \multicolumn{2}{c}{HM-EQA} & \multicolumn{2}{c}{MT-HM3D} & \multicolumn{2}{c}{EXPRESS-Bench} & \multicolumn{2}{c}{A-EQA (184)} \\
\cmidrule(lr){2-3} \cmidrule(lr){4-5} \cmidrule(lr){6-7} \cmidrule(lr){8-9}
 & SR & Steps (↓) 
 & SR & Steps (↓)
 & LLM Score & $E_{path}$ (↑)
 & LLM-Match & $E_{path}$ (↑) \\
\midrule
GPT-4V (OpenEQA) & --   & --   & --     & --   & --    & --      & 41.8 & 7.5 \\
Explore-EQA & 58.4 & 0.52 & 36.2*  & 0.64 & --    & --      & 46.9* & 23.4 \\
Graph-EQA & 63.5 & \textbf{0.20} & 45.63* & 0.45 & --    & --      & 30.1*^\dagger  & -- \\
Memory-EQA & 63.4  & 0.40  & \textbf{55.1}   & \textbf{0.41} & --    & --      & 36.8^\dagger  & -- \\ 
Fine-EQA & 56.0 & 0.54 & --     & --   & 63.95 & 25.58 & 43.3^\dagger & 29.2 \\
3D-Mem & --   & --   & --     & --   & --    & --      & \textbf{52.6} & \textbf{42.0} \\
\midrule
FAST-EQA ($k=3$) & \makecell{$\mathbf{69.2}$ \\ $\pm 0.7$} 
         & \makecell{$0.65$ \\ $\pm 0.01$} 
         & \makecell{$50.5$ \\ $\pm 0.3$} 
         & \makecell{$0.52$ \\ $\pm 0.01$} 
         & \makecell{$\mathbf{68.7}$ \\ $\pm 0.5$} 
         & \makecell{$\mathbf{29.25}$ \\ $\pm 0.55$} 
         & \makecell{$49.0$ \\ $\pm 1.7$} 
         & \makecell{$27.70$ \\ $\pm 1.70$} \\
\midrule
FAST-EQA ($k=5$) & \makecell{$\mathbf{71.0}$ \\ $\pm 0.7$} 
         & \makecell{$0.65$ \\ $\pm 0.01$} 
         & \makecell{$52.36$ \\ $\pm 0.3$} 
         & \makecell{$0.52$ \\ $\pm 0.01$} 
         & \makecell{$\mathbf{68.8}$ \\ $\pm 0.5$} 
         & \makecell{$\mathbf{27.41}$ \\ $\pm 0.55$} 
         & \makecell{$46.73$ \\ $\pm 1.7$} 
         & \makecell{$31.93$ \\ $\pm 1.70$} \\
\bottomrule
\end{tabularx}
}
\end{table*}

\section{Parameter Tuning}
When retrieving relevant memory, we use a tunable parameter $\lambda$ to weigh Prismatic and CLIP scores. This combination scoring allows the agent to retrieve observations that align with both the focused target goal and the question-answering goal. We vary the value of parameter $\lambda$ and measure the question-answering success rate and normalized steps taken in \ref{tab:lamda}. Based on this, we choose $\lambda = 0.7$ in our final experiments.

\begin{table}
    \centering
    \begin{tabular}{lcc}
    \toprule
       $\lambda$ & SR \\
    \midrule
       0.0 & 69.0 \\
       0.5 & 76.0 \\
       0.6 & 72.0 \\
       0.7 & 78.0 \\
       1.0 & 72.0 \\
    \bottomrule
    \end{tabular}
    \caption{Tuning the scoring parameter $\lambda$ for relevant memory retrieval on a fixed subset of HM-EQA}
    \label{tab:lamda}
\end{table}

\section{Prompts}

\begin{figure*}[!t]
    \centering
    \includegraphics[width=\textwidth]{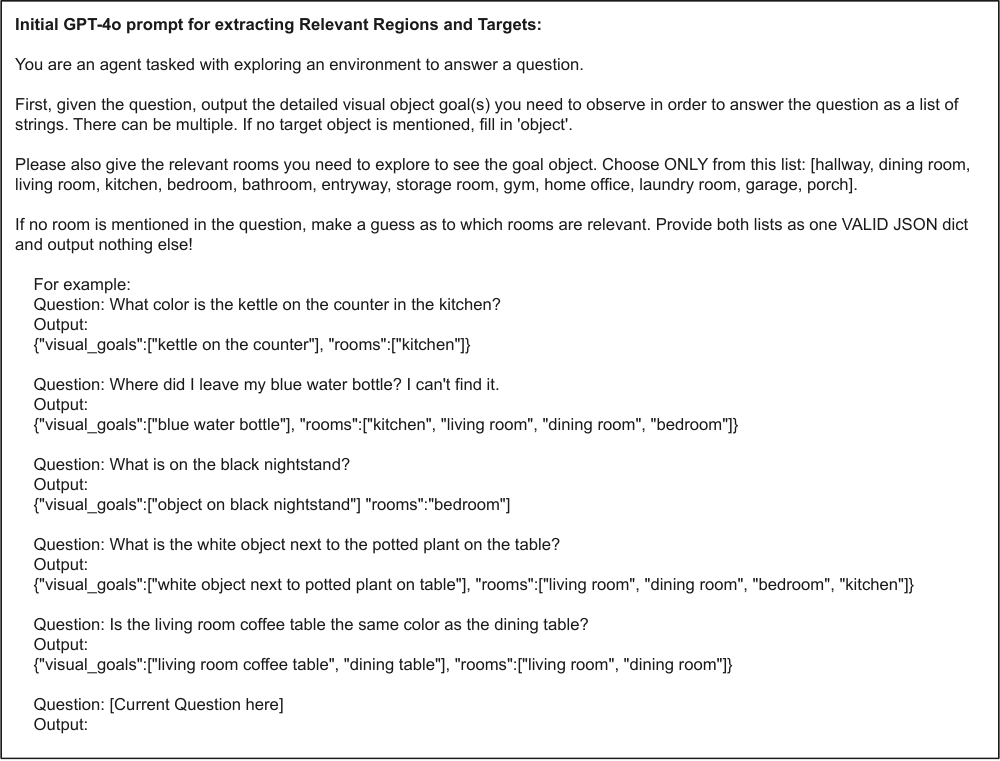}
    \caption{Prompt for extracting relevant regions and targets}
    \label{fig:prompt_1}
\end{figure*}

FAST-EQA employs a variety of prompts when interacting with VLMs or LLMs across different stages of the pipeline. As illustrated in Figure \ref{fig:prompt_1}, the prompt shown is issued at the beginning of each episode to identify the relevant regions and visual targets corresponding to the given question. We provide illustrative few-shot examples to clarify what is meant by relevant regions and visual targets for different types of questions. The model’s response is returned in the form of a JSON string, which is subsequently parsed to generate a structured list of relevant regions and visual targets.

\begin{figure*}[!t]
    \centering
    \includegraphics[width=\textwidth]{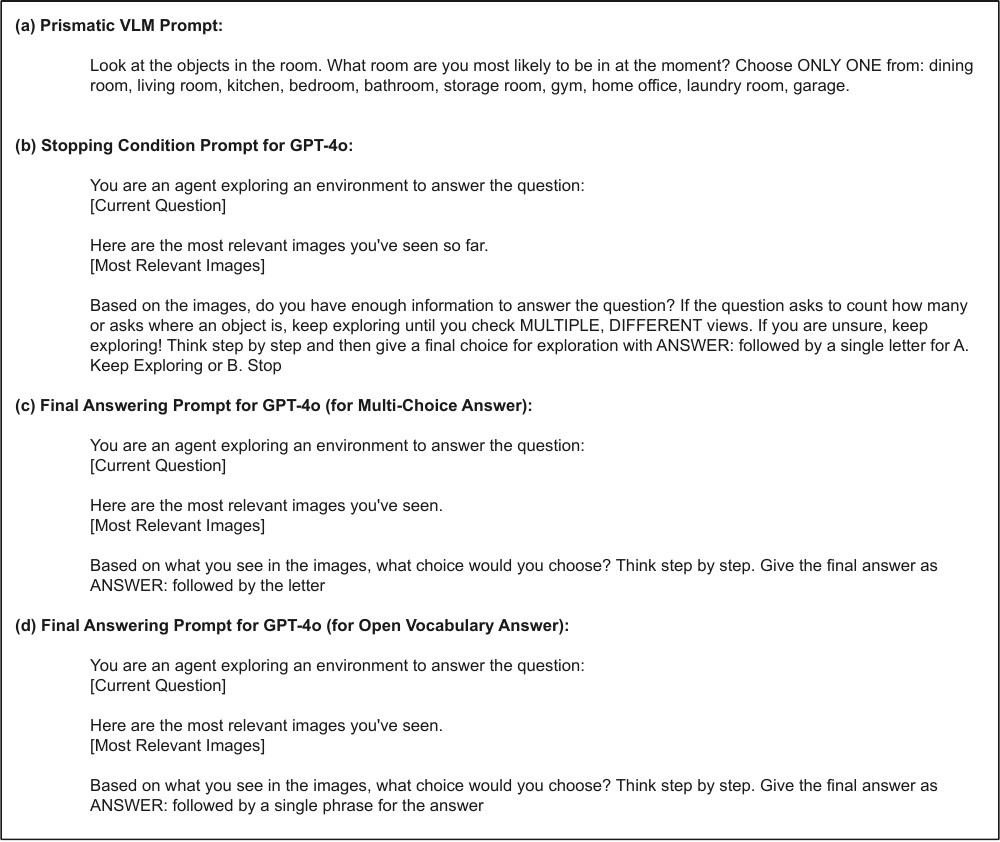}
    \caption{Prompts for (a) determining current region $R_{t}$, (b) stopping condition, (c) final answering for MCQA, and (d) final answering for Open Vocabulary Questions}
    \label{fig:prompt_2}
\end{figure*}

To determine the current region $R_{t}$, FAST-EQA queries Prismatic-VLM \cite{karamcheti2024prismatic} with the current observation $o_{t}$ using the prompt shown in Figure~\ref{fig:prompt_2} (a). For the stopping condition, FAST-EQA queries GPT-4o \cite{hurst2024gpt} with the prompt in Figure \ref{fig:prompt_2} (b) which includes the question $Q$ and the most relevant images retrieved from visual memory, asking whether sufficient information is available to answer the question. An exception is made for questions involving counting or object existence, where GPT-4o is instructed to continue exploring, as such tasks often require a more exhaustive examination of the scene.

For final question answering, FAST-EQA uses the prompt in Figure \ref{fig:prompt_2} (c) for multi-choice QA and (d) for open vocabulary QA. The prompt includes instruction to think step-by-step to encourage chain-of-thought reasoning. 

\bibliographystyle{ieeenat_fullname}
\bibliography{main}